\documentclass[runningheads]{llncs}

\usepackage[utf8]{inputenc}
\usepackage[T1]{fontenc}
\usepackage{graphicx}
\usepackage{amsmath,amssymb}
\usepackage{booktabs}
\usepackage{multirow}
\usepackage{hyperref}
\usepackage{caption}
\usepackage{subcaption}
\usepackage{xcolor}
\usepackage{tabularx}   
\usepackage{array}
\usepackage{makecell}   

\newcommand{\hdr}[1]{\multicolumn{1}{c}{\scriptsize\makecell{#1}}}

\newcommand{\maseheader}{%
Dataset & $L$ & \hdr{Na\"ive} & \hdr{Seas.\\Na\"ive} & \hdr{ARIMA}
& \hdr{Auto-\\former} & \hdr{DLinear} & \hdr{Patch\\TST}
& \hdr{Chronos} & \hdr{\textbf{Ours}} \\}

\graphicspath{{./}{./figures/}}

\begin{document}

\title{Parameter-Efficient Adaptation of Pretrained Language Models
for Time-Series Forecasting}

\author{
    Tamanna Kumavat\inst{1} \and
    Georg Brunner\inst{2} \and
    Kyriakos Flouris\inst{3}
}
\institute{
    University of Zurich, Switzerland\\
    \email{tamannasurendar.kumavat@uzh.ch}
    \and
    Computer Vision Laboratory, ETH Zurich, Switzerland\\
    \email{brunnerg@student.ethz.ch}
    \and
    MRC Biostatistics Unit, University of Cambridge, United Kingdom\\
    \email{kyriakos.flouris@mrc-bsu.cam.ac.uk}
}
\authorrunning{T. Kumavat et al.}
\titlerunning{Parameter-Efficient LM Adaptation for Time-Series Forecasting}

\maketitle

\begin{abstract}
We study the adaptation of pretrained language models to univariate time-series forecasting through a parameter-efficient transfer learning framework, with the goal of understanding which design choices drive effective cross-modal transfer. While language models operate on discrete textual tokens, time series consist of continuous numerical observations with temporal dependencies. To bridge this modality gap, we project fixed-length time-series patches directly into the embedding space of a pretrained GPT-2 backbone, bypassing textual tokenization and treating the Transformer as a generic sequence encoder. Through controlled ablation studies on seven benchmark datasets spanning energy, weather, traffic, and finance, we analyze the effects of (i)~representation strategy (continuous embeddings versus textual serialisation), (ii)~adaptation regime (frozen backbone versus partial or full fine-tuning), (iii)~architectural components such as adapters, pooling strategies, and prediction heads, and (iv)~input context length. Continuous patch-based embeddings consistently outperform textual prompting and randomly initialised backbones. The adapted pipeline attains MASE within the range of specialised forecasting architectures while updating less than 1\% of total model parameters. Results further indicate that freezing the pretrained backbone and training lightweight projection and adapter modules provides a favourable accuracy--efficiency trade-off with stable behaviour across varying context lengths.

\end{abstract}

\section{Introduction}

Time-series forecasting is a fundamental problem in machine learning with applications spanning energy systems, finance, transportation, and meteorology. Classical approaches such as ARIMA and exponential smoothing~\cite{box2015time,hyndman2008} rely on linearity and stationarity assumptions frequently violated in complex real-world systems. Deep learning methods based on recurrent and convolutional architectures~\cite{hochreiter1997} capture nonlinear temporal dependencies, and Transformer-based architectures~\cite{vaswani2017attention,zhou2021informer,wu2021autoformer,nie2023patchtst} have achieved strong benchmark performance. However, these models are typically trained from scratch, requiring substantial labelled data and computational resources.

Large language models (LLMs) such as GPT~\cite{radford2019,brown2020} have demonstrated remarkable transfer capabilities across natural language tasks. This has motivated growing interest in repurposing pretrained Transformers for time-series forecasting~\cite{gruver2023,ansari2024chronos,garza2023timegpt,jin2024timellm,zhou2023onefitsall}. Yet adapting language models to numerical forecasting introduces a fundamental modality gap~\cite{tan2024llms,ye2024survey}: natural language consists of discrete tokens with semantic structure, whereas time series comprise continuous-valued measurements with temporal dependence and scale sensitivity.

Existing representation strategies span textual serialisation~\cite{gruver2023}, discretisation into finite vocabularies~\cite{ansari2024chronos}, and direct projection into embedding space~\cite{nie2023patchtst}. The importance of such representation choices is well established in generative modelling, where learning efficient low-dimensional latent embeddings has been shown to yield more compact and faithful data descriptions than operating through entangled or redundant representations~\cite{flouris2023canonical,flouris2024explicit}. Nevertheless, performance varies substantially across these time-series approaches depending on dataset characteristics, training regimes, and evaluation settings, making it difficult to disentangle architectural effects from representational choices~\cite{qiu2024tfb,revisitingllms2025,flouris2025localized}.

Rather than introducing a new architecture or pursuing state-of-the-art
performance, this work develops a systematic
adaptation and evaluation pipeline to isolate three central factors:
(i)~the inductive biases of self-attention as a general sequence-processing
mechanism, (ii)~the representational prior induced by language-model
pretraining, and (iii)~the contribution of parameter-efficient task
alignment modules. We select GPT-2~\cite{radford2019} as the pretrained
backbone for its causal (decoder-only) architecture, which naturally
aligns with temporal forecasting where each patch should attend only
to preceding context. Preliminary experiments indicated that this
causal structure outperformed bidirectional (BERT) and encoder-decoder
(T5) alternatives in our pipeline, consistent with the autoregressive
nature of the forecasting task. GPT-2's open availability and
widespread adoption in related work further enable controlled
ablation studies under realistic computational budgets.

Our contributions are as follows:
(1)~A reproducible benchmarking framework for evaluating pretrained
Transformer-based forecasting across datasets, horizons, and context lengths.
(2)~Comprehensive ablation studies analysing representation strategies,
adaptation schemes, architectural components, and data scale.
(3)~An empirical comparison between pretrained and randomly initialised
backbones, isolating the effect of language-model pretraining.
(4)~Evaluation of cross-domain zero-shot generalisation.
(5)~Practical guidelines for parameter-efficient adaptation in
resource-constrained settings.

Code, experimental configurations, and instructions for reproducing all
tables and figures are available at
\url{https://github.com/tamannaKumavat/GPT-TS}.

\section{Related Work}

\textbf{Transformers for time series.}\quad
Transformer architectures have been widely applied to time-series forecasting. Informer~\cite{zhou2021informer} introduces ProbSparse attention for long-sequence efficiency, while Autoformer~\cite{wu2021autoformer} incorporates series decomposition with auto-correlation. PatchTST~\cite{nie2023patchtst} segments time series into patches and processes them as token sequences, achieving strong performance with channel independence. DLinear~\cite{zeng2023dlinear} challenges the necessity of attention by demonstrating competitive results with simple linear layers. Our work builds on the patching strategy of PatchTST but replaces the randomly initialised backbone with a pretrained language model, enabling the study of cross-modal transfer.

\textbf{Language models for time series.}\quad
Several approaches repurpose pretrained LLMs for temporal forecasting. Gruver et al.~\cite{gruver2023} serialise numerical values as text and perform zero-shot forecasting through next-token prediction. Chronos~\cite{ansari2024chronos} discretizes time-series values into a finite vocabulary and pretrains a T5-based model on large-scale temporal corpora. TimeGPT~\cite{garza2023timegpt} and Time-LLM~\cite{jin2024timellm} propose reprogramming strategies to align time-series inputs with language model representations. One Fits All~\cite{zhou2023onefitsall} freezes a pretrained GPT-2 backbone and trains lightweight heads for multiple temporal tasks. Our approach shares the frozen-backbone principle with One Fits All but differs in the systematic ablation of representation strategies, adaptation regimes, and architectural components, providing controlled evidence for design choices that prior work evaluates only implicitly.

\textbf{Parameter-efficient transfer learning.}\quad
Adapter modules~\cite{houlsby2019parameter} and low-rank adaptation (LoRA)~\cite{hu2022lora} enable task-specific alignment of pretrained models while updating only a small fraction of parameters, mitigating catastrophic forgetting~\cite{mccloskey1989catastrophic} and overfitting on small downstream datasets. These techniques have been extensively validated in NLP but remain underexplored for cross-modal transfer to time series. We employ bottleneck adapters following Houlsby et al.~\cite{houlsby2019parameter} and systematically evaluate their capacity--accuracy trade-off in the forecasting setting.

\section{Method}

\begin{figure}[t]
    \centering
    \includegraphics[width=\textwidth]{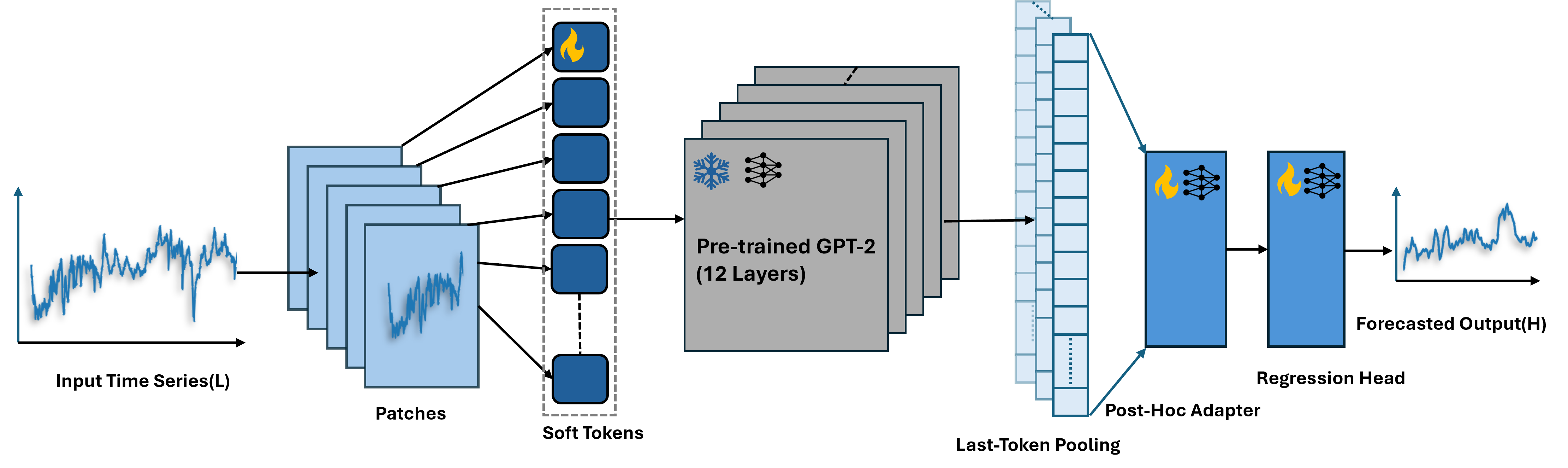}
    \caption{\textbf{Architecture overview.} A context window of
    length $L$ is standardised and segmented into non-overlapping
    patches. Each patch is linearly projected into the embedding
    space of a frozen GPT-2 backbone and processed as a sequence
    of continuous ``soft tokens.'' A lightweight residual adapter
    refines hidden states before last-token pooling and projection
    to an $H$-step forecast.}
    \label{fig:pipeline}
\end{figure}

The proposed pipeline (Figure~\ref{fig:pipeline}) adapts a pretrained
GPT-2 backbone for time-series forecasting through three stages:
patch-based embedding, Transformer processing, and direct multi-step
forecast projection.

Given a univariate time series $\{x_t\}_{t=1}^{N}$, the data is split
chronologically into training and test segments. Each observation is
standardised via z-score normalisation using mean and variance computed
exclusively on the training segment: $x'_t = (x_t - \mu)/\sigma$.
A sliding window of context length $L$ and forecast horizon $H$
produces input--target pairs for supervised training. The standardised
context window $\mathbf{x} \in \mathbb{R}^{L}$ is then partitioned
into $K = \lfloor (L - P)/S \rfloor + 1$ non-overlapping patches of
length $P{=}8$ with stride $S{=}8$, following~\cite{nie2023patchtst}.
Patching aggregates local temporal dynamics into compact
representations and reduces the effective sequence length from $L$ to
$K$, lowering self-attention cost from $\mathcal{O}(L^2)$ to
$\mathcal{O}((L/P)^2)$.

Each patch $\mathbf{p}_k \in \mathbb{R}^{P}$ is projected into the
Transformer embedding dimension $d$ via a learned linear
transformation followed by layer normalisation:
$\mathbf{e}_k = \text{LayerNorm}(\mathbf{W}_p \mathbf{p}_k +
\mathbf{b}_p)$, where $\mathbf{W}_p \in \mathbb{R}^{d \times P}$.
The resulting sequence
$\mathbf{E} = [\mathbf{e}_0, \ldots, \mathbf{e}_{K-1}]$ forms
continuous ``soft tokens'' that reside directly in the latent space
of the pretrained model, bypassing textual tokenization entirely via
the \texttt{inputs\_embeds} interface. We employ GPT-2 Small
(12 layers, 768-dimensional hidden states, 124M
parameters)~\cite{radford2019} as the backbone, which processes
the embedded patches and produces hidden representations
$\mathbf{H} = \text{GPT-2}(\mathbf{E}) \in \mathbb{R}^{K \times d}$.

A residual adapter~\cite{houlsby2019parameter} refines the hidden states for task-specific
alignment: $\tilde{\mathbf{H}}_k = \mathbf{H}_k +
\text{Adapter}(\mathbf{H}_k)$, where the adapter is a bottleneck MLP
projecting $d \to \lfloor d/r \rfloor \to d$ with GELU activation
and default ratio $r{=}8$. Last-token pooling selects
$\mathbf{z} = \tilde{\mathbf{H}}_{K-1}$, naturally aligned with
GPT-2's causal left-to-right processing where the final token
summarizes all preceding context through self-attention. The pooled
representation is mapped to the $H$-step forecast via
$\hat{\mathbf{y}} = \mathbf{W}_o \mathbf{z} + \mathbf{b}_o$,
following a direct multi-step (DMS) strategy that predicts all future
timesteps simultaneously, avoiding error accumulation from recursive
decoding.

In the default configuration, the GPT-2 backbone is frozen and only
the patch projection ($\mathbf{W}_p, \mathbf{b}_p$), layer
normalisation, adapter, and regression head
($\mathbf{W}_o, \mathbf{b}_o$) are trained, amounting to
approximately 197K out of 124M total parameters (0.16\%). The model is optimized by minimizing mean
squared error on the standardised scale:
\begin{equation}
\mathcal{L} = \frac{1}{BH}\sum_{i=1}^{B}\sum_{h=1}^{H}
\bigl(y^{(i)}_{t+h} - \hat{y}^{(i)}_{t+h}\bigr)^2,
\end{equation}
where $B$ is the mini-batch size. All reported metrics are computed
after inverse-transforming predictions to the original data scale.

\section{Experimental Setup}

We evaluate on seven benchmark datasets spanning energy systems
(ETTh1, ETTh2, ETTm1, ETTm2~\cite{zhou2021informer}), power
consumption (Electricity), transportation (Traffic), and meteorology
(Weather)~\cite{wu2021autoformer}. Each dataset is used for
univariate forecasting with a single target variable. Although several
datasets are originally multivariate, we focus on the univariate
setting to isolate representation and transfer effects. This
channel-independent strategy, also adopted by
PatchTST~\cite{nie2023patchtst}, has been shown to act as an
implicit regulariser that can prevent overfitting in
Transformer-based forecasters. Datasets are
split chronologically (80/20 train/test), with a 10\% validation subset
extracted from the training tail for model selection. To ensure
computational feasibility across the full grid of configurations, each
dataset is restricted to a fixed chronological prefix (see
Appendix~\ref{app:full_results}, Table~\ref{tab:datasets} for exact
lengths) that preserves temporal ordering and distributional properties;
all methods---including baselines---are trained and evaluated on
identical truncated segments to ensure fair comparison. Complete dataset
statistics, target variables, and seasonal periods are reported in
Appendix~\ref{app:full_results}.

Context lengths $L \in \{96, 192, 336\}$ and forecast horizons
$H \in \{24, 48, 96\}$ follow standard benchmarking
conventions~\cite{nie2023patchtst,wu2021autoformer}. All benchmark
results are computed under a boundary forecasting protocol: each model
is trained exclusively on the training segment and produces a single
direct $H$-step forecast at the train--test boundary using the final
$L$ observations as input. Forecasting accuracy is assessed using
MASE as the primary metric, defined as the ratio of forecast MAE to
the in-sample seasonal na\"ive MAE:
\begin{equation}
\text{MASE} = \frac{\frac{1}{H}\sum_{h=1}^{H}
|y_{t+h} - \hat{y}_{t+h}|}
{\frac{1}{N_{\text{train}}-m}\sum_{t=m+1}^{N_{\text{train}}}
|y_t - y_{t-m}|},
\end{equation}
where $m$ is the seasonal period. Values below 1.0 indicate
improvement over the seasonal na\"ive baseline. MAE and MSE are also
reported in Appendix~\ref{app:full_results}.

We compare against classical baselines (Na\"ive last-value repeat,
Seasonal Na\"ive, and ARIMA$(5,1,0)$ via
Statsmodels), modern neural architectures
(Autoformer~\cite{wu2021autoformer},
DLinear~\cite{zeng2023dlinear}, and
PatchTST~\cite{nie2023patchtst} implemented via
NeuralForecast), and the Chronos
foundation model~\cite{ansari2024chronos} using
\texttt{chronos-t5-small} with the sample mean of 20 forecast draws
as the point prediction. All neural models share identical optimisation controls and training
budgets to ensure fair comparisons. Detailed hyperparameters are
reported in Appendix~\ref{app:training}. Random seeds are fixed to
42 across all experiments.

All entries for the proposed pipeline use the linear prediction head and adapter ratio $r{=}8$, held fixed across every dataset and horizon rather than tuned per dataset. Section~\ref{sec:results} ablations show that an MLP head and $r{=}2$ reduce MASE further; the table entries therefore reflect a single fixed configuration, not the lowest achievable error.

\section{Results}
\label{sec:results}

We present the full baseline comparison at the representative horizon $H{=}48$, followed by ablation studies analysing representation design, transfer learning, adaptation regimes, architectural sensitivity, and generalisation. Complete results for $H \in \{24, 96\}$ across all datasets and context lengths are provided in Appendix~\ref{app:full_results}.

\begin{table}[t]
\centering
\caption{MASE results for $H{=}48$ across all datasets and context
lengths. Best per row in \textbf{bold}. Classical baselines are
context-independent. The final row reports the average rank across
the 21 configurations at this horizon (lower is better).}
\label{tab:mase_h48}
\footnotesize
\setlength{\tabcolsep}{4pt}
\renewcommand{\arraystretch}{1.05}
\begin{tabular}{@{}lcrrrrrrrr@{}}
\toprule
\maseheader
\midrule
ETTh1 & 96 & 0.388 & 1.009 & 0.393 & 0.780 & 0.443 & 0.710 & 0.431 & \textbf{0.363} \\
ETTh1 & 192 & 0.388 & 1.009 & 0.393 & 1.875 & 0.508 & 0.471 & 0.420 & \textbf{0.386} \\
ETTh1 & 336 & 0.388 & 1.009 & 0.393 & 0.766 & \textbf{0.280} & 2.115 & 1.190 & 0.385 \\
\addlinespace
ETTh2 & 96 & 2.158 & \textbf{1.493} & 1.894 & 1.989 & 1.946 & 1.502 & 1.985 & 1.989 \\
ETTh2 & 192 & 2.158 & 1.493 & 1.894 & 3.004 & 1.326 & \textbf{1.137} & 1.394 & 2.003 \\
ETTh2 & 336 & 2.158 & 1.493 & 1.894 & 2.720 & \textbf{1.228} & 1.550 & 1.714 & 2.067 \\
\addlinespace
ETTm1 & 96 & 0.983 & 0.624 & 0.991 & 0.336 & 0.235 & 0.222 & 1.460 & \textbf{0.202} \\
ETTm1 & 192 & 0.983 & 0.624 & 0.991 & 1.805 & 0.348 & \textbf{0.212} & 1.189 & 0.229 \\
ETTm1 & 336 & 0.983 & 0.624 & 0.991 & 0.508 & 0.476 & \textbf{0.436} & 0.657 & 0.839 \\
\addlinespace
ETTm2 & 96 & 1.808 & 0.844 & 1.155 & 2.302 & 0.500 & 0.741 & \textbf{0.237} & 0.359 \\
ETTm2 & 192 & 1.808 & 0.844 & 1.155 & 1.268 & 0.528 & 0.567 & 0.303 & \textbf{0.219} \\
ETTm2 & 336 & 1.808 & 0.844 & 1.155 & 4.350 & 0.828 & 0.693 & 0.726 & \textbf{0.352} \\
\addlinespace
Electricity & 96 & 0.170 & 0.098 & 0.170 & \textbf{0.092} & 0.097 & 0.101 & 0.125 & 0.169 \\
Electricity & 192 & 0.170 & 0.098 & 0.170 & 0.170 & \textbf{0.094} & 0.096 & 0.162 & 0.152 \\
Electricity & 336 & 0.170 & \textbf{0.098} & 0.170 & 0.288 & 0.102 & 0.106 & 0.134 & 0.152 \\
\addlinespace
Traffic & 96 & 4.166 & 1.041 & 3.639 & 0.921 & 0.848 & \textbf{0.641} & 0.860 & 0.938 \\
Traffic & 192 & 4.166 & 1.041 & 3.639 & 0.861 & 0.602 & 0.601 & \textbf{0.518} & 1.035 \\
Traffic & 336 & 4.166 & 1.041 & 3.639 & 0.831 & 0.637 & \textbf{0.634} & 1.034 & 1.253 \\
\addlinespace
Weather & 96 & 1.349 & 1.342 & 1.574 & 1.129 & 1.213 & 1.163 & 1.336 & \textbf{1.100} \\
Weather & 192 & 1.349 & 1.342 & 1.574 & 1.335 & 1.464 & 1.199 & \textbf{0.998} & 1.259 \\
Weather & 336 & 1.349 & 1.342 & 1.574 & \textbf{0.757} & 1.356 & 1.203 & 1.552 & 1.260 \\
\midrule
\multicolumn{2}{@{}l}{Avg.\ rank} & 6.29\,(8) & 4.52\,(5) & 6.10\,(7) & 5.45\,(6) & 3.14\,(2) & \textbf{2.81\,(1)} & 4.19\,(4) & 3.50\,(3) \\
\bottomrule
\end{tabular}
\end{table}

\begin{figure}[t]
    \centering
    \includegraphics[width=0.85\textwidth]{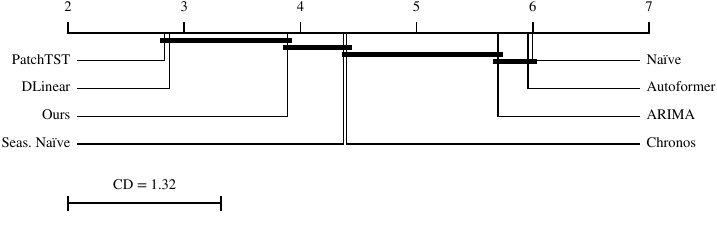}
    \caption{Critical-difference diagram over 63 dataset$\times$context$\times$horizon configurations ($k{=}8$, $N{=}63$, $\alpha{=}0.05$). Methods joined by a horizontal bar are not statistically separable (Nemenyi CD${=}1.32$).}
    \label{fig:cd_diagram}
\end{figure}

\textbf{Baseline comparison.}\quad
Table~\ref{tab:mase_h48} reports MASE at $H{=}48$; complete results at $H{=}24$ and $H{=}96$ appear in Appendix~\ref{app:full_results}. The purpose of this comparison is to establish that the pipeline operates in a performance regime comparable to established forecasters: ablations conducted on a testbed far off the pace would not generalise. Averaged over all 63 dataset--context--horizon configurations, the pipeline attains a mean rank of $3.89$ against $2.83$ for PatchTST, $2.87$ for DLinear, and $4.40$ for Chronos (Friedman $\chi^2(7){=}123.2$, $p{<}10^{-22}$). The Nemenyi critical difference at $\alpha{=}0.05$ is $1.32$~\cite{demsar2006}, so the pipeline is not statistically separable from PatchTST, DLinear, Chronos, or the Seasonal Na\"ive baseline; it is separated from ARIMA, Autoformer, and the Na\"ive baseline (Figure~\ref{fig:cd_diagram}).\footnote{Ranks are computed treating each (dataset, context length, horizon) as an independent configuration ($N{=}63$). Classical baselines (Na\"ive, Seasonal Na\"ive, ARIMA) are context-independent, so their three $L$ rows per (dataset, horizon) pair are not fully independent replicates; the pooled test therefore mildly overstates the effective sample size for those methods. Restricting to one row per (dataset, horizon) pair ($N{=}21$) yields the same ordering.}

Per-dataset behaviour varies. Performance is weaker on ETTh2, where MASE consistently exceeds 1.0 for all neural methods, suggesting this dataset favours models with stronger linear inductive biases such as DLinear. On Electricity, lightweight models (Seasonal Na\"ive, DLinear) attain the lowest MASE, indicating that the additional capacity of a pretrained Transformer provides limited benefit on highly structured periodic signals. On Traffic, PatchTST and Chronos rank first at most context lengths. An observation across all three horizons is that the pipeline's MASE does not degrade at $L{=}336$, in contrast to Autoformer, which exhibits substantial instability at longer contexts (e.g., MASE of $4.350$ on ETTm2 at $L{=}336$).

\begin{figure}[htbp]
    \centering
    \begin{subfigure}[b]{0.49\textwidth}
        \centering
        \includegraphics[width=\linewidth]{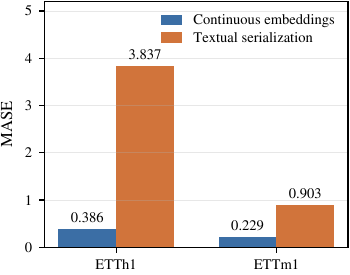}
        \caption{Soft vs.\ hard prompting}
        \label{fig:soft_hard}
    \end{subfigure}
    \hfill
    \begin{subfigure}[b]{0.49\textwidth}
        \centering
        \includegraphics[width=\linewidth]{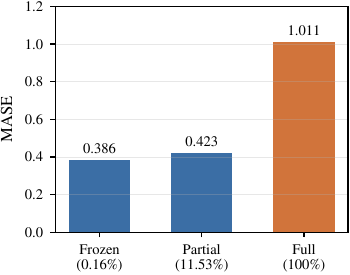}
        \caption{Backbone adaptation regime}
        \label{fig:freezing}
    \end{subfigure}
    \\[0.6em]

    \begin{subfigure}[b]{0.49\textwidth}
        \centering
        \includegraphics[width=\linewidth]{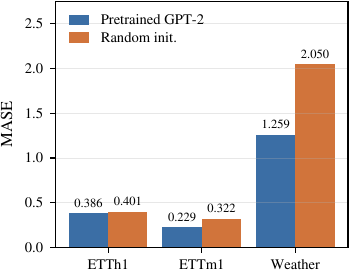}
        \caption{Pretrained vs.\ random init.}
        \label{fig:pretraining}
    \end{subfigure}
    \hfill
    \begin{subfigure}[b]{0.49\textwidth}
        \centering
        \includegraphics[width=\linewidth]{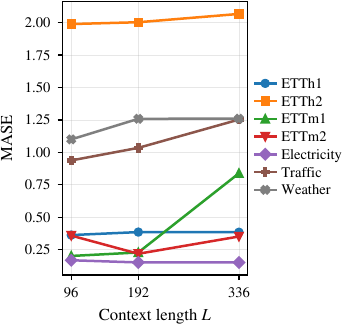}
        \caption{Context length scaling}
        \label{fig:context}
    \end{subfigure}
    \\[0.6em]

    \begin{subfigure}[b]{0.49\textwidth}
        \centering
        \includegraphics[width=\linewidth]{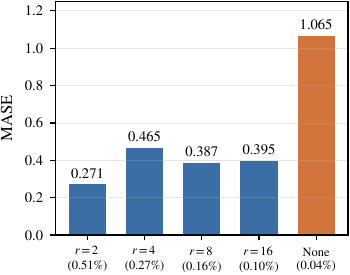}
        \caption{Adapter capacity ($r$)}
        \label{fig:adapter}
    \end{subfigure}
    \hfill
    \begin{subfigure}[b]{0.49\textwidth}
        \centering
        \includegraphics[width=\linewidth]{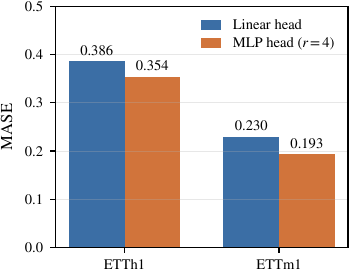}
        \caption{Linear vs.\ MLP head}
        \label{fig:head}
    \end{subfigure}

    \caption{\textbf{Ablation studies.} (a)~Continuous embeddings dramatically outperform textual serialisation. (b)~Frozen backbone with $<$1\% trainable parameters achieves best accuracy; full fine-tuning degrades performance. (c)~Pretrained GPT-2 consistently outperforms random initialization under identical conditions. (d)~Context length scaling is stable with no catastrophic degradation. (e)~Adapter bottleneck ratio $r{=}2$ yields best accuracy; removing adapters degrades substantially. (f)~MLP prediction head improves over linear across datasets. All results at $H{=}48$, $L{=}192$ unless noted.}
    \label{fig:ablations}
\end{figure}

\textbf{Soft tokens vs.\ hard prompting.}\quad
Figure~\ref{fig:ablations}a compares the proposed continuous embedding injection against textual serialisation where numerical values are converted to strings and processed via GPT-2's standard tokenizer. The soft-token interface substantially and consistently outperforms hard prompting across all datasets at $H{=}48$. On ETTh1, soft tokens achieve MASE$=0.386$ versus $3.84$ for hard prompting---nearly an order-of-magnitude difference. On ETTm1, the gap is fourfold ($0.229$ vs.\ $0.90$). The training truncation rate was zero for both, confirming the gap stems from representational inefficiency rather than context overflow. Textual serialisation expands individual values into multiple BPE tokens, exhausting the context window, and forces the model to operate through embeddings pretrained on linguistic rather than quantitative structure. Continuous patch embeddings preserve magnitude relationships while controlling token count.

\textbf{Backbone adaptation regime.}\quad
Figure~\ref{fig:ablations}b compares three regimes on ETTh1 at $H{=}48$: frozen backbone, partially frozen (last two Transformer blocks unfrozen), and full fine-tuning. The frozen regime achieves the best accuracy (MASE$=0.386$) while training only 0.16\% of parameters. Unfreezing the last two blocks increases trainable parameters to 11.53\% but degrades performance (MASE$=0.423$). Full fine-tuning ($100\%$ parameters) results in substantially worse accuracy (MASE$=1.011$), consistent with overfitting or degradation of pretrained representations~\cite{mccloskey1989catastrophic}. These results indicate that the pipeline benefits from reusing frozen pretrained representations while adapting only lightweight modules.

\textbf{Effect of language-model pretraining.}\quad
Figure~\ref{fig:ablations}c compares the pretrained GPT-2 backbone against an identically structured randomly initialised model under matched conditions (frozen backbone, $L{=}192$, $H{=}48$). Pretraining consistently improves performance across all datasets: on ETTh1 MASE decreases from $0.401$ to $0.386$, on ETTm1 from $0.322$ to $0.229$, and on Weather from $2.050$ to $1.259$. Because the backbone remains frozen, these gains cannot be attributed to optimisation dynamics and are instead consistent with the hypothesis that large-scale language pretraining induces reusable sequence-processing representations that transfer to numerical domains.

\textbf{Context length scaling.}\quad
Figure~\ref{fig:ablations}d shows MASE across context lengths $L \in \{96, 192, 336\}$ at $H{=}48$. On ETTm2, performance improves monotonically as additional history is provided. On ETTh1 and Weather, improvements saturate beyond $L{=}192$ with marginal variation at $L{=}336$. Crucially, no dataset exhibits severe degradation at larger contexts, contrasting with certain baselines that show instability under long input windows. This stability derives from the patch-based interface, which reduces effective sequence length from $L$ to $K \approx L/P$, controlling quadratic attention cost.

\textbf{Adapter capacity.}\quad
Figure~\ref{fig:ablations}e varies the bottleneck ratio $r \in \{2,4,8,16\}$ under a frozen backbone. Removing the adapter degrades MASE to $1.065$, confirming that lightweight task-specific adaptation is critical. The best performance is obtained at $r{=}2$ (MASE$=0.271$, $0.51\%$ trainable parameters), with increasing $r$ progressively degrading accuracy ($r{=}8$: $0.386$; $r{=}16$: $0.395$). This reveals a clear accuracy--capacity trade-off: sufficient adapter width is necessary for stable alignment between pretrained representations and the forecasting objective.

\textbf{Architectural sensitivity.}\quad
Figure~\ref{fig:ablations}f compares linear and MLP prediction heads. The MLP head consistently improves performance: on ETTh1, MASE drops from $0.386$ to $0.354$; on ETTm1, from $0.230$ to $0.193$. The parameter overhead is modest (from 0.16\% to 0.25\% of total). Additionally, last-token pooling consistently outperforms mean pooling, with MASE improving on ETTh1 from $1.277$ (mean) to $0.386$ (last-token), consistent with GPT-2's causal processing structure, where the final token integrates all preceding context.

\begin{figure}[t]
    \centering
    \begin{subfigure}[b]{0.49\textwidth}
        \centering
        \includegraphics[width=\linewidth]{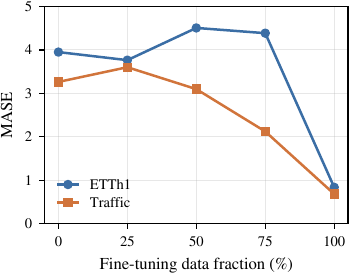}
        \caption{Training data fraction effect}
        \label{fig:data_pct}
    \end{subfigure}
    \hfill
    \begin{subfigure}[b]{0.49\textwidth}
        \centering
        \includegraphics[width=\linewidth]{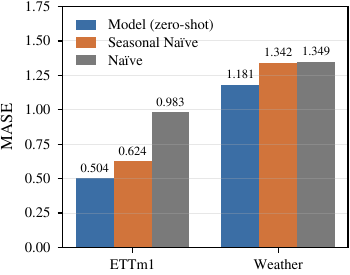}
        \caption{Zero-shot cross-domain transfer}
        \label{fig:cross_domain}
    \end{subfigure}
    \caption{\textbf{Data efficiency and generalisation.} (a)~Performance under reduced training data; behavior is dataset-dependent with Traffic scaling smoothly and ETTh1 requiring full supervision. (b)~Zero-shot cross-domain transfer consistently outperforms na\"ive baselines, with ${\sim}19\%$ improvement on ETTm1 and ${\sim}12\%$ on Weather over Seasonal Na\"ive.}
    \label{fig:generalisation}
\end{figure}

\textbf{Data efficiency and cross-domain generalisation.}\quad
Figure~\ref{fig:generalisation}a shows performance under subsampled training data. On Traffic, MASE decreases substantially with more supervision, dropping from $3.26$ (0\%) to $0.682$ (100\%), though with non-monotonic behaviour at low fractions. On ETTh1, performance remains poor at reduced fractions (MASE$\approx 3.8$--$4.5$) with a sharp improvement only at 100\% ($0.837$), suggesting this dataset requires sufficient supervision for reliable generalisation. Figure~\ref{fig:generalisation}b evaluates zero-shot cross-domain transfer: a model trained on one source dataset is evaluated on unseen targets without fine-tuning. On ETTm1, the model achieves MASE$=0.504$, improving over the seasonal na\"ive baseline ($0.624$) by approximately $19\%$. On Weather, zero-shot MASE$=1.181$ outperforms seasonal na\"ive ($1.342$) by ${\sim}12\%$, demonstrating that the pretrained backbone and learned forecasting interface can capture reusable temporal structure across domains.

\section{Discussion and Conclusion}

This work provides a systematic empirical analysis of the design choices underlying parameter-efficient adaptation of pretrained language models for time-series forecasting. Rather than proposing a new state-of-the-art architecture, our primary contribution is a set of controlled ablation studies that isolate the effects of representation strategy, adaptation regime, and architectural components. Several key findings emerge.

First, \emph{representation design is the dominant factor} in cross-modal adaptation. Continuous patch-based embeddings substantially outperform textual serialisation by avoiding tokenization inefficiency, embedding misalignment, and formatting sensitivity. These results suggest that embedding-level alignment appears substantially more effective than textual serialisation in our experiments and may be necessary for stable time-series adaptation.

Second, \emph{language-model pretraining provides transferable structural benefits} that extend beyond text. The pretrained GPT-2 backbone consistently outperforms an identically structured randomly initialised model even when frozen. These results are consistent with the hypothesis that pretraining induces reusable sequence-processing inductive biases (long-range dependency modelling, hierarchical structure, and multi-scale relationships via self-attention) rather than transferring linguistic semantics. An alternative explanation, that better-conditioned weight matrices lead to more stable gradient flow, cannot be fully ruled out and merits further investigation through representation probing.

Third, \emph{parameter-efficient adaptation outperforms full fine-tuning} in this setting. Freezing the backbone and training lightweight modules ($<$1\% of parameters) achieves superior accuracy compared to partial or full fine-tuning. The adapter ablation confirms that some task-specific alignment is necessary, but lightweight residual refinement suffices.

Fourth, \emph{patch-based segmentation enables stable context scaling} without the catastrophic degradation observed in some Transformer baselines. Architectural interface choices (MLP head, last-token pooling) produce secondary but meaningful improvements.

Together, these findings yield practical guidelines for
resource-constrained deployment: use continuous patch embeddings
rather than textual serialisation, freeze the pretrained backbone,
attach a lightweight adapter ($r \leq 4$) with an MLP prediction
head and last-token pooling; this configuration achieves
competitive accuracy while training less than 1\% of parameters.

Several limitations warrant discussion. The study focuses on
univariate forecasting with a single backbone (GPT-2 Small);
larger models or multivariate settings may exhibit different
dynamics. Data efficiency is dataset-dependent, and no single
configuration universally dominates. Extending the evaluation
to additional test configurations and seeds would provide
stronger statistical grounding. While preliminary experiments favoured the causal GPT-2
architecture over bidirectional and encoder-decoder alternatives,
a systematic multi-backbone comparison with full reporting remains
future work.

The most substantive question opened by our results is \emph{what, specifically, transfers from language pretraining to numerical forecasting}. Section~\ref{sec:results} shows that a frozen pretrained backbone consistently outperforms a randomly initialised one under identical conditions, but two hypotheses are compatible with this observation: pretraining induces reusable sequence-processing inductive biases (long-range dependency modelling, hierarchical structure), or pretraining simply yields better-conditioned weight matrices that support more stable forward passes. Distinguishing them is the natural extension of this work into a full research paper. Concrete probes include layer-wise Centered Kernel Alignment~\cite{kornblith2019similarity} between pretrained and random-init backbones on time-series inputs, attention-pattern comparison across the two initialisations, and a weight-spectrum control that rescales the random-init backbone to match the singular-value distribution of the pretrained one, isolating the conditioning hypothesis. Complementary directions include extending the framework to multivariate forecasting, investigating model-scale effects, and combining language pretraining with time-series-specific self-supervised objectives.

\bibliographystyle{splncs04}
\bibliography{references}

\newpage
\appendix

\section{Dataset Configuration}
\label{app:full_results}

Table~\ref{tab:datasets} summarizes dataset-specific details
including target variables, sampling frequencies, seasonal periods
used for MASE computation, and chronological prefix lengths used
for computational feasibility.

\begin{table}[h]
\centering
\caption{Dataset configuration. The seasonal period $m$ defines
the scaling denominator in MASE. All datasets use patch length
$P{=}8$ and stride $S{=}8$.}
\label{tab:datasets}
\footnotesize
\setlength{\tabcolsep}{6pt}
\renewcommand{\arraystretch}{1.05}
\begin{tabular}{@{}llcrr@{}}
\toprule
Dataset & Target & Frequency & $m$ & Max length \\
\midrule
ETTh1 & OT & Hourly & 24 & 5{,}000 \\
ETTh2 & OT & Hourly & 24 & 5{,}000 \\
ETTm1 & OT & 15-min & 96 & 8{,}000 \\
ETTm2 & OT & 15-min & 96 & 8{,}000 \\
Electricity & Col.\ 0 & Hourly & 24 & 8{,}000 \\
Traffic & Col.\ 2 & Hourly & 24 & 5{,}000 \\
Weather & T ($^\circ$C) & Hourly & 24 & 10{,}000 \\
\bottomrule
\end{tabular}
\end{table}

\section{Training and Hyperparameter Details}
\label{app:training}

All neural models (the proposed method and baselines) are trained
under identical optimisation controls. The proposed GPT-2
soft-token forecaster uses AdamW with learning rate
$2 \times 10^{-4}$ and weight decay $10^{-2}$, batch size 8
with gradient accumulation 2 (effective batch size 16), for up
to 15 epochs with early stopping (patience 3) monitored on
validation MAE. Gradient norms are clipped at 1.0 and training
uses mixed precision (FP16). The backbone is frozen by default;
trainable components include the patch projection layer, layer
normalisation, residual adapter (bottleneck ratio $r{=}8$, GELU
activation), and a linear prediction head. This amounts to less
than 1\% of total model parameters. The benchmark results in
Table~\ref{tab:mase_h48} use this conservative configuration;
ablation studies in the main text show that an MLP head and
$r{=}2$ yield further improvements.

Autoformer~\cite{wu2021autoformer},
DLinear~\cite{zeng2023dlinear}, and
PatchTST~\cite{nie2023patchtst} are trained via
NeuralForecast with early stopping (patience 3) and validation
checks every 200 steps. ARIMA uses a fixed $(5,1,0)$
configuration via Statsmodels. Chronos~\cite{ansari2024chronos}
uses \texttt{chronos-t5-small} with the sample mean of 20
forecast draws as the point prediction. Na\"ive and Seasonal
Na\"ive baselines are deterministic and require no training.

\section{Results for Additional Forecast Horizons}
\label{app:additional_horizons}

Table~\ref{tab:mase_h24} presents MASE results for short-horizon
forecasting ($H{=}24$). Rank ordering is consistent with the $H{=}48$
results in Table~\ref{tab:mase_h48}: PatchTST ranks first (mean rank $2.29$),
followed by DLinear (mean rank $3.29$) and the proposed pipeline (mean rank $3.86$).

\begin{table}[h]
\centering
\caption{MASE results for horizon $H{=}24$. Best per row in
\textbf{bold}. The final row reports the average rank across the
21 configurations at this horizon (lower is better).}
\label{tab:mase_h24}
\footnotesize
\setlength{\tabcolsep}{4pt}
\renewcommand{\arraystretch}{1.05}
\begin{tabular}{@{}lcrrrrrrrr@{}}
\toprule
\maseheader
\midrule
ETTh1 & 96 & 0.353 & 0.853 & 0.344 & 1.788 & 0.349 & 0.361 & 0.485 & \textbf{0.276} \\
ETTh1 & 192 & 0.353 & 0.853 & 0.344 & 1.321 & \textbf{0.216} & 0.315 & 0.505 & 0.268 \\
ETTh1 & 336 & 0.353 & 0.853 & 0.344 & 1.103 & \textbf{0.267} & 0.829 & 0.881 & 0.420 \\
\addlinespace
ETTh2 & 96 & 2.245 & 1.570 & 1.973 & 1.961 & 2.024 & \textbf{1.414} & 1.892 & 1.968 \\
ETTh2 & 192 & 2.245 & 1.570 & 1.973 & 2.943 & 1.361 & \textbf{1.322} & 1.426 & 2.005 \\
ETTh2 & 336 & 2.245 & 1.570 & 1.973 & 2.297 & \textbf{1.199} & 1.467 & 1.670 & 2.042 \\
\addlinespace
ETTm1 & 96 & 0.650 & 0.670 & 0.656 & 0.938 & 0.231 & \textbf{0.161} & 0.934 & 0.182 \\
ETTm1 & 192 & 0.650 & 0.670 & 0.656 & 0.378 & 0.357 & \textbf{0.227} & 0.791 & 0.332 \\
ETTm1 & 336 & 0.650 & 0.670 & 0.656 & 1.793 & 0.435 & \textbf{0.269} & 0.495 & 0.533 \\
\addlinespace
ETTm2 & 96 & 0.933 & 0.681 & 0.452 & 2.691 & \textbf{0.267} & 0.585 & 0.270 & 0.592 \\
ETTm2 & 192 & 0.933 & 0.681 & 0.452 & 0.684 & 0.347 & 0.547 & 0.277 & \textbf{0.168} \\
ETTm2 & 336 & 0.933 & 0.681 & 0.452 & 1.493 & 0.527 & 0.579 & 0.666 & \textbf{0.254} \\
\addlinespace
Electricity & 96 & 0.237 & 0.142 & 0.237 & 0.174 & 0.112 & \textbf{0.104} & 0.127 & 0.202 \\
Electricity & 192 & 0.237 & 0.142 & 0.237 & 0.195 & 0.126 & \textbf{0.120} & 0.160 & 0.211 \\
Electricity & 336 & 0.237 & 0.142 & 0.237 & 0.252 & 0.147 & \textbf{0.130} & 0.137 & 0.216 \\
\addlinespace
Traffic & 96 & 4.227 & 1.138 & 3.692 & 1.372 & 0.919 & \textbf{0.724} & 0.846 & 1.158 \\
Traffic & 192 & 4.227 & 1.138 & 3.692 & 1.005 & 0.811 & 0.706 & \textbf{0.524} & 1.063 \\
Traffic & 336 & 4.227 & 1.138 & 3.692 & 0.996 & 0.784 & \textbf{0.737} & 1.149 & 0.947 \\
\addlinespace
Weather & 96 & \textbf{0.304} & 0.353 & 0.365 & 0.544 & 0.733 & 0.284 & 0.313 & 0.426 \\
Weather & 192 & 0.304 & 0.353 & 0.365 & 1.159 & 1.192 & 0.371 & 0.438 & \textbf{0.278} \\
Weather & 336 & \textbf{0.304} & 0.353 & 0.365 & 0.957 & 0.689 & 0.325 & 0.676 & 0.459 \\
\midrule
\multicolumn{2}{@{}l}{Avg.\ rank} & 5.74\,(7) & 4.86\,(5) & 5.02\,(6) & 6.71\,(8) & 3.29\,(2) & \textbf{2.29\,(1)} & 4.24\,(4) & 3.86\,(3) \\
\bottomrule
\end{tabular}
\end{table}

Table~\ref{tab:mase_h96} presents MASE results for long-horizon
forecasting ($H{=}96$). At this horizon DLinear ranks first (mean rank $2.17$),
followed by PatchTST (mean rank $3.38$) and the Seasonal Na\"ive baseline (mean rank $3.71$);
the proposed pipeline ranks fourth ($4.31$), reflecting the increasing difficulty of longer horizons for adaptation-based approaches.

\begin{table}[h]
\centering
\caption{MASE results for horizon $H{=}96$. Best per row in
\textbf{bold}. The final row reports the average rank across the
21 configurations at this horizon (lower is better).}
\label{tab:mase_h96}
\footnotesize
\setlength{\tabcolsep}{4pt}
\renewcommand{\arraystretch}{1.05}
\begin{tabular}{@{}lcrrrrrrrr@{}}
\toprule
\maseheader
\midrule
ETTh1 & 96 & 0.553 & 1.247 & 0.564 & 1.702 & 0.704 & 1.878 & 0.941 & \textbf{0.465} \\
ETTh1 & 192 & 0.553 & 1.247 & 0.564 & 1.533 & 0.902 & 1.118 & 1.117 & \textbf{0.486} \\
ETTh1 & 336 & 0.553 & 1.247 & 0.564 & 1.885 & \textbf{0.482} & 1.559 & 1.264 & 0.576 \\
\addlinespace
ETTh2 & 96 & 1.984 & 1.340 & 1.755 & 1.938 & 1.752 & \textbf{1.306} & 1.965 & 2.313 \\
ETTh2 & 192 & 1.984 & 1.340 & 1.755 & 2.923 & \textbf{1.286} & 1.597 & 1.661 & 2.225 \\
ETTh2 & 336 & 1.984 & 1.340 & 1.755 & 2.873 & \textbf{1.173} & 1.507 & 1.622 & 2.287 \\
\addlinespace
ETTm1 & 96 & 1.720 & 1.033 & 1.729 & 3.299 & 0.827 & \textbf{0.774} & 2.733 & 1.176 \\
ETTm1 & 192 & 1.720 & 1.033 & 1.729 & 5.734 & \textbf{0.933} & 1.211 & 1.982 & 1.113 \\
ETTm1 & 336 & 1.720 & 1.033 & 1.729 & 1.341 & 0.985 & \textbf{0.932} & 1.563 & 1.406 \\
\addlinespace
ETTm2 & 96 & 3.295 & 1.953 & 2.552 & 1.875 & \textbf{1.780} & 1.836 & 2.920 & 1.780 \\
ETTm2 & 192 & 3.295 & 1.953 & 2.552 & 3.986 & 1.809 & 1.955 & 1.876 & \textbf{1.648} \\
ETTm2 & 336 & 3.295 & 1.953 & 2.552 & \textbf{1.351} & 1.977 & 1.879 & 1.770 & 1.859 \\
\addlinespace
Electricity & 96 & 0.149 & \textbf{0.078} & 0.149 & 0.103 & 0.082 & 0.133 & 0.128 & 0.164 \\
Electricity & 192 & 0.149 & \textbf{0.078} & 0.149 & 0.124 & 0.086 & 0.105 & 0.179 & 0.138 \\
Electricity & 336 & 0.149 & \textbf{0.078} & 0.149 & 0.353 & 0.081 & 0.129 & 0.138 & 0.150 \\
\addlinespace
Traffic & 96 & 4.511 & 1.020 & 3.968 & 1.306 & 0.851 & \textbf{0.730} & 0.964 & 1.145 \\
Traffic & 192 & 4.511 & 1.020 & 3.968 & 1.143 & \textbf{0.731} & 0.773 & 0.921 & 0.956 \\
Traffic & 336 & 4.511 & 1.020 & 3.968 & 0.853 & \textbf{0.682} & 0.777 & 1.012 & 1.089 \\
\addlinespace
Weather & 96 & 2.829 & 2.793 & 3.135 & 3.131 & \textbf{2.345} & 2.535 & 2.909 & 2.513 \\
Weather & 192 & 2.829 & 2.793 & 3.135 & 2.432 & 2.700 & 2.641 & \textbf{2.358} & 2.974 \\
Weather & 336 & 2.829 & 2.793 & 3.135 & 2.265 & 2.148 & 2.214 & 2.297 & \textbf{1.574} \\
\midrule
\multicolumn{2}{@{}l}{Avg.\ rank} & 5.98\,(8) & 3.71\,(3) & 5.98\,(7) & 5.71\,(6) & \textbf{2.17\,(1)} & 3.38\,(2) & 4.76\,(5) & 4.31\,(4) \\
\bottomrule
\end{tabular}
\end{table}

\end{document}